\documentclass[11pt,a4paper]{article}
\usepackage[hyperref]{acl2020}
\usepackage{times}
\usepackage{latexsym}

\usepackage{microtype}
\usepackage[caption=false, font=footnotesize]{subfig}
\usepackage{amssymb}
\usepackage{amsmath}
\usepackage{graphicx}
\usepackage{bm}

\aclfinalcopy 
\def\aclpaperid{***} 

\title{Model Reduction of Shallow CNN Model for Reliable  Deployment of Information Extraction from Medical Reports}

\author{Abhishek K Dubey \and Alina Peluso \and Jacob Hinkle  \\ Oak Ridge National Lab, Oak Ridge \AND
        Devanshu Agarawal \\  University of Tennessee, Knoxville \And Zilong  Tan \\ Carnegie Mellon University,  Pittsburgh}

\date{}

\begin{document}
\maketitle
\begin{abstract}
{\small
Shallow Convolution Neural Network (CNN) is a time-tested tool for the information extraction 
from cancer pathology reports. 
Shallow CNN performs competitively on this task to other deep learning models including BERT, which 
holds the state-of-the-art for many NLP tasks.
The main insight behind this eccentric phenomenon is that the information extraction from cancer 
pathology reports require only a small number of domain-specific text segments to perform 
the task, thus making the most of the texts and contexts excessive for the task.
Shallow CNN model is well-suited to identify these key short text segments 
from the labeled training set; however, the identified text segments remain obscure to humans.
In this study, we fill this gap by developing a model reduction tool to make a reliable connection between 
CNN filters and relevant text segments by discarding the spurious connections. 
We reduce the complexity of shallow CNN representation by approximating it with a linear transformation 
of n-gram presence representation with a non-negativity and sparsity prior on the transformation weights
to obtain an interpretable model.  
Our approach bridge the gap between the conventionally perceived tradeoff boundary 
between accuracy on the one side and explainability on the other by model reduction.
}

\end{abstract}

\section{Introduction}

The recent advancement in data acquisition and computing technologies has spurred the growth in the 
application of artificial intelligence in the medical domain. 
These advances have a significant impact on low-level detection and recognition tasks in texts and images.
There are a set of risks and challenges when we apply these technologies to mission-critical 
applications such as health care. 
For reliable deployment of AI to health care, we must probe the model's 
understanding of the task and ensure that it is not merely exploiting the systemic artifacts 
embedded in data.
Model distillation becomes essential in this context towards achieving this goal.
For this reason, there has been increased research interest in model distillation or developing techniques for explaining the decision process of deep learning models.
This work falls under the first category.

In this paper, we conduct a direct analysis of CNN representations 
by mapping it to the n-gram features of text reports. 
We do so by modeling a CNN representation $\boldsymbol{\phi}(\mathbf{x})$ 
of a text report $\mathbf{x}$  as a linear function 
$\boldsymbol{\phi}(\mathbf{x})=\mathbf{A}\,tp(\mathbf{x})$
of n-gram features $tp(\mathbf{x})$.
A common hypothesis is that CNN representations collapse irrelevant information in the
texts and thus $\boldsymbol{\phi}$ should not be uniquely invertible.
Also, for the information extraction task, we hypothesis that most of the information in the
reports is irrelevant, thus $\boldsymbol{\phi}$ should only act on a small 
proportion of the text segments.
Hence, we pose the problem of model reduction as finding a sparse model $\mathbf{A}$ and by doing 
so we obtained insights into which text segments maybe that are important for the 
prediction.

Our contributions are as follows. First, we propose a general method to approximate
CNN representation for text. 
We discuss and evaluate non-negativity constraint penalty as a model prior to 
get a sparse interpretable model.
Second, we show that despite using a linear map of n-gram representation 
$\mathbf{A}\,tp(\mathbf{x})$  for approximation, our method achieves the same accuracy as 
the CNN model.
Third, we apply pseudo-inverse of the linear map $\mathbf{A}^{\dagger}$ to approximate the sample 
reconstructions of a text reports from a given representation for explaining an individual prediction.
The rest of manuscript is organized as follows.
Section \ref{sec:relatedwork} provides an overview of related work. 
Section \ref{sec:method} describes document representations and our method. 
Section \ref{sec:dataset} describes de-identified datasets used for the experiments.
Section \ref{sec:results} and \ref{sec:conclusion} presents the experimental results and discussion.

\section{Related work}
\label{sec:relatedwork}
Several techniques have been proposed to help improve our understanding of Neural network representations~\citep{erhan2009visualizing,
mahendran2015understanding,yosinski2015understanding,nguyen2016synthesizing, shwartz2017opening, doshi2017towards}.
\citet{bach2015pixel} proposed the layer-wise relevance propagation (LRP) technique to explain 
the DNN classification decisions. LRP redistributes predictions backward through layers 
of the model using local redistribution rules until it assigns a relevance score to input components.
A second technique, sensitivity analysis~\citep{baehrens2010explain, simonyan2013deep}, explains model predictions 
based on the model’s gradients at input locations.
\citet{ribeiro2016should} proposed a generalized technique, LIME, that provides locally 
interpretable explanations of model predictions by utilizing the samples within proximity of the input samples. 
They perturb the input in the proximity and see how the model behave and then
weight the perturations by their proximity to the original input to learn an locally interpretable 
model.
\citet{zhou2016learning} recently proposed the Class Activation Mapping (CAM) algorithm 
using the global average pooling before the final output layer for identifying the all discriminative regions 
used by CNN models for a particular class. 
\citet{selvaraju2017grad} proposed a generalized CAM (Grad-CAM), which fuses
the class-conditional CAM with existing pixel-space gradient visualization techniques
to find fine-grained detailed discriminative regions. 
In this work, instead of developing methods to explain the prediction of individual predictions,
we approximate CNN representations as a linear transformation of n-gram presence features  with a 
non-negativity and sparsity prior on the model weights to build a reduced explainable model.  

\section{Method}
\label{sec:method}
\begin{figure}
\centering
\includegraphics[width=0.55\textwidth]{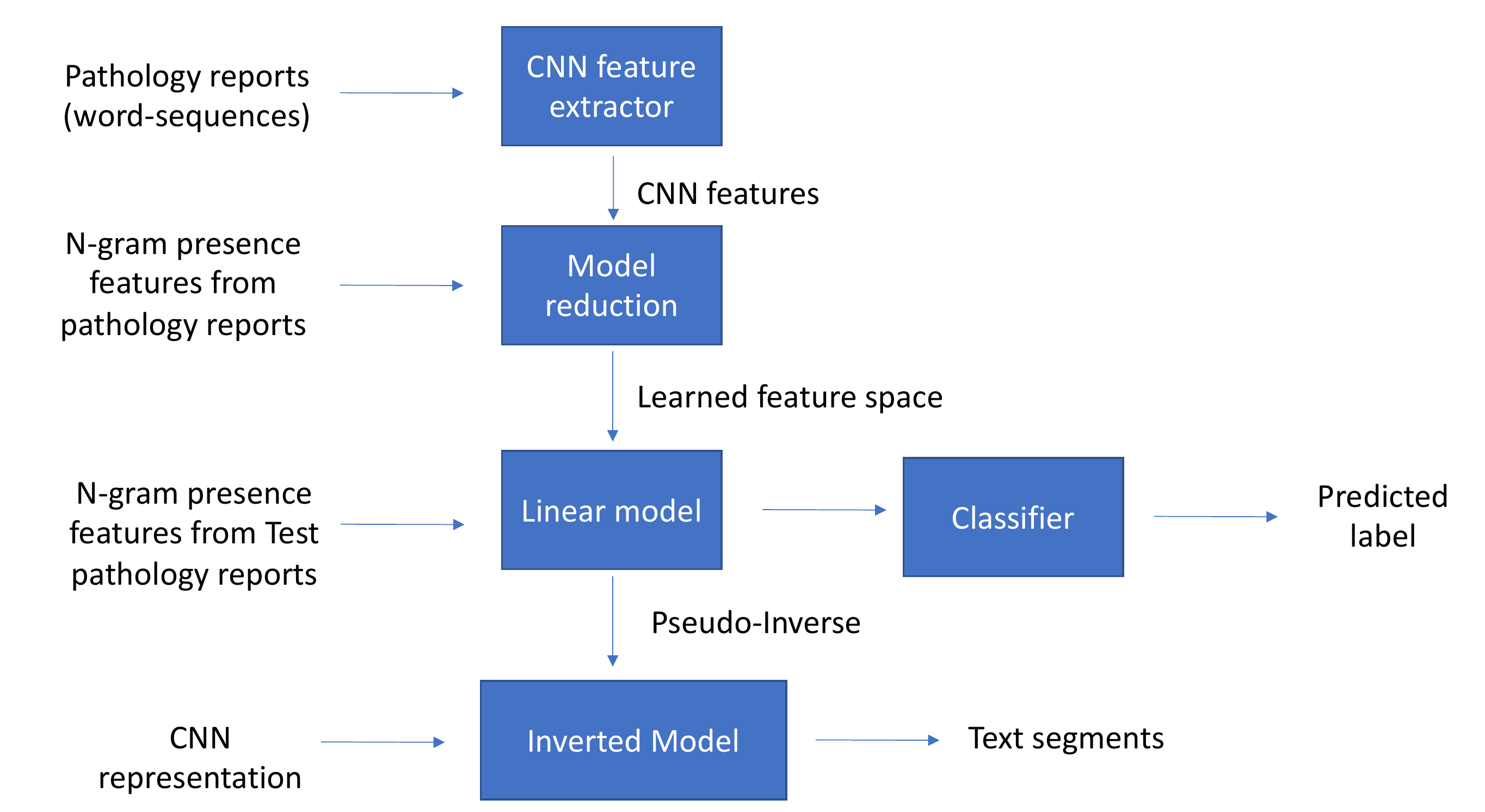}
\caption{{\small A workflow diagram of the model reduction framework.}}
\label{fig:model-reduction-framework}
\end{figure}

This section introduces our model reduction framework. 
We illustrate the complete workflow in
Figure~\ref{fig:model-reduction-framework}.
First, we extract the CNN representations of the text reports
using the shallow CNN network~\cite{kim2014convolutional,devanshu2019dkl}, also described 
in Section~\ref{subsec:cnn-network}.  
Second, we generate the n-gram presence representations of the text reports,
which is introduced in Section~\ref{subsec:ngram-presence-representation}.
Third, we construct the linear map between the CNN representations of the text reports
and n-gram presence representations and discard the spurious  connections to 
obtain a sparse interpretable linear model 
as described in Section~\ref{subsec:linear-approx}.
Next, we review CNN representations and n-gram presence representations 
of the text reports and  describe the model reduction method in details.
\subsection{CNN representation}
\label{subsec:cnn-network}
Shallow CNN architecture 
is first developed
for text classification by \cite{kim2014convolutional} and later successfully applied
to cancer pathology reports by \cite{qiu2017deep}.
The CNN takes a text document $\mathbf{x}$ as its input, which is represented as a 
sequence of word tokens $x_i$ from a provided vocabulary of size $V$.
The input sequence is first passed through a word embedding layer that maps each token to a word vector
of dimension $E$, where $E<V$. 
The embedding layer maps  the document $\mathbf{x}$ to a matrix $\mathbf{H}$ whose elements are given by
\begin{equation}
h_{ij} = \operatorname{embed}(\mathbf{x})|_{i,j}
= e_{x_i j},
\end{equation}
where $e_{x_i j}$ is the word embedding vector of work token $x_i$.
The embedding output is then passed through $P \cdot F$ convolutional filters, $P$ is number of  convolution modules
and $F$ is the number of convolutional filter per module.
A convolutional module with $F$ convolutional filters all of width $M$ are defined as
\begin{equation}
\displaystyle
\phi_j(\mathbf{x}) = \max_i \operatorname{\sigma}\left(\sum_{m,n=1}^{M,E} h_{i+m-1,n} w_{mnj}  + b_j\right),
\end{equation}
where $w_{mnj}$ and $b_j$ are elements of the $M\times E\times F$ convolutional weight matrix and $F$-dimensional bias vector $\mathbf{b}$ respectively, 
and $\operatorname{\sigma}(\cdot)$ is the rectified linear activation function.
A maxpool layer is used at the end of convolution operation to render a translational invariant 
feature.
%
%
The output vectors of all convolutional modules are then concatenated together, producing a CNN  representation of 
dimension $P \cdot F$.
%

\subsection{N-gram presence representation}
\label{subsec:ngram-presence-representation}
$N$-gram representation is a popular bag of word representation of the text  documents
that capture short sequence information and ignores longer ones. 
Term-frequency and inverse-document frequency (tf-idf) is two most
popular choice for document representation in this category~\citep{ramos2003using}.
In this work, we rather choose to use $n$-gram presence representation to represent  the
cancer reports  for the information extraction task.
The motivation behind this choice is to select a representation that is least susceptible
to produce overfitted model. 
In $n$-gram presence representation, 
We  use the presence or absence of the short word sequences in the reports
as a feature for the representation. 
We denote the n-gram presence features of a document $d$ by 
\begin{equation}
tp(d) = [tp(t_{1},d)...tp(t_{p},d)]^{T}
\end{equation}
where $tp(t_{i}, d)=1$ when n-gram $t_{i}$ is present in the document $d$ and $tp(t_{i},d)=0$ when 
$t_{i}$ is absent.
%
%
%

\subsection{Model reduction}
\label{subsec:linear-approx}
In this section, we introduce our method to distill the cnn representations 
by finding the active set of n-gram presence features in the pathology reports
via non-negative linear map approximation of the cnn representations. 
Consider the functional dependence of the label variable $Y \in \mathcal{R}$
on the explanatory document variables $\mathbf{X}$ through a non-linear mapping
$Y=\boldsymbol{\phi}(\mathbf{X})$.
Given a set of observations $\{(\mathbf{x}_i, y_i)\}_{i=1}^{n}$ of $\mathbf{X}$ and $Y$,
we estimate the function $\boldsymbol{\phi}$ through training a shallow CNN model as in \cite{kim2014convolutional}
and obtain a set of CNN representations $\{\boldsymbol{\phi}(\mathbf{x_{i}})\}_{i=1}^{n}$.
We also extract a set of n-gram presence features $\{tp(\mathbf{x}_{i})\}_{i=1}^{n}$ 
from the pathology reports.
We find the approximation of the CNN representations with a linear map of the n-gram presence features
with non-negative weights
by solving a non-negative least squares problem,
\begin{eqnarray}
 \label{eq:nnls}
 \arg\min_{\mathbf{A}} \sum_{i=1}^{n} \| (\mathbf{A} \, tp(\mathbf{x}_i) - \boldsymbol{\phi}(\mathbf{x}_i))\|^{2}, \hspace{2pt} A_{ij} > 0.
\end{eqnarray}
By minimizing~\ref{eq:nnls}, we obtain a dense linear map with non-negative weights.
To discard the spurious associations, we suppress the values of $A_{ij}$ to zero of
n-gram features that contributes little. 
Overall contribution of  the $i$th n-gram feature $tp(.)_{i}$  to the cnn representation 
is calculated by aggregating weights of the $i$th column of the linear map $\sum_{j=1}^{P.F} A_{ij}$.
We keep only the top contributing n-gram presence features for building a reduced interpretable 
model.
We note that our decision to put a non-negativity prior on the weights of the linear map is 
in order to discard the complex solutions that may overfit when $p>n$ where $p$ is the 
number n-gram features and $n$ is available text reports. 
%


\section{Dataset}
\label{sec:dataset}
To validate our model,
we used primary-site extraction task  on the de-identified pathology reports,
which were collected from $5$ SEER cancer registries (CT, HI, KY, NM, Seattle).
The reports were labeled by state cancer registries as per coding guidelines issued by SEER.
The primary-site extraction task is to label the report with the standard ICD-O-3 topography codes 
that is used to encode the tumor location.
A summary of the labels including the ICD-O-3 code 
and an example of unstructured text section of the 
pathology report are provided by ~\citet{Dubey2019mrmr-lsir}.
The cancer pathology reports were provided in XML format.
We discarded the meta-data associated with the pathology reports
and used only unstructured text sections of the XML file
for the task.
We adopted the same pre-processing steps as in
~\citep{Dubey2019mrmr-lsir}.
%
%
%
%


\section{Experimental results}
\label{sec:results}
\begin{table*}[tb]
  \centering
  {\small
  \begin{tabular}{ccccccc}
  \hline
  \multicolumn{1}{c}{Features} & \multicolumn{1}{c}{Reduction} & Classifier & Model type & Active features & Accuracy\\
  \hline
  N-gram Term-Freq. & LSIR~\cite{dubey2019inverse} & kNN & dense & 6855 & 0.80\\
  N-gram Term-Freq. & mRMR+LSIR~\cite{Dubey2019mrmr-lsir} & kNN & sparse & 2000 & 0.82\\
  \hline
  CNN (our impl.) & - & Softmax Linear & dense & - & 0.87\\
  N-gram presence & model reduction (Section~\ref{subsec:linear-approx}) & SIGP & sparse & 3000 & 0.87\\
  \hline
 \end{tabular}
 }
  \caption{
      {\small
      Average accuracy scores
      for the tumor primary-site extraction task of $4$ different models 
      on the test set obtained by tenfold cross-validation partition of a  corpus 
      of $825$ de-identified reports,  which consists of $6$ prevalent class-labels 
      each category containing at least $50$ samples. The results of traditional ML methods 
      are taken from the precursor studies~\citep{dubey2019inverse,Dubey2019mrmr-lsir} for comparison.}}
  \label{table:classifcation-comparison}
\end{table*}

The goal of our experiments is to compare the classification accuracy and complexity of the reduced model
with the CNN baseline and two other inverse-regression based methods~\citep{Dubey2019mrmr-lsir,dubey2019inverse}
explored previously for information extraction from cancer pathology reports.
We report the average test accuracy on ten test folds obtained by a ten-fold stratified cross-validation 
of the de-identified reports.
For  each  of  the training-test split, we further split the training fold into training and validation sets, 
stratified  to  maintain  relative  class  frequencies.  

For the CNN baseline, we used CNN model by~\citep{kim2014convolutional} with 
$300$ word embedding dimensions, $3$ convolutional modules of filter-width 3, 4 and 5, $128$ convolution filters
per modules and l2-regularization parameter of 0.001 in the loss function.
The feature network weights and classifier weights of the shallow CNN model are initialized 
randomly.
The CNN model is trained in each fold for up to $100$ epochs using the Adam optimizer with minibatch size $32$ 
and a learning rate of $0.001$. We used early stopping with patience $10$.
We used the CNN model that gives the highest validation accuracy score 
for prediction as the final model.
We used the final models for prediction on test sets to estimate the average classification accuracy.
We also passed all the de-identified reports through the final model for each fold to obtain
the CNN representations for model reduction. 

For the model reduction experiments, we extracted $n$-gram features
of length 3, 4, and 5 from the reports using CountVectorizer function from the sklearn library. 
We obtain the corresponding 
n-gram presence features of de-identified reports for all ten folds by using the same seed
as the CNN experiment.
We discard any n-gram features that have a document frequency strictly lower than $3$.
We extracted a total of $33814$ features. We binarize the features to obtain the $n$-gram presence 
features (see Section~\ref{subsec:ngram-presence-representation}). 
We used a fast non-negative least squares solver by~\cite{kim2013non} to obtained the linear map $\mathbf{A}$. 
We used the default parameter settings recommended in the released package~\cite{kiSrDh12}. 
We computed the overall contribution of the features to the CNN features
by summing their contributions. 
We keep only the top $3K$ features for building a reduced model.
We build a classifier using subspace-induced gaussian processes (SIGP)~\cite{tan2018subspace}
to get the final prediction for each fold. We used the radial basis function (RBF) kernel in the SIGP 
and learned its scale parameter through bayesian optimization.

Table~\ref{table:classifcation-comparison} shows a comparison of the classification accuracy of
the reduced model with other methods for the primary-site classification task.
%
%
The results show that the reduced model achieves the same classification accuracy 
as the CNN model despite only using less than $10\%$ of the total n-gram features. 
Results also show a gap in the accuracy between the traditional inverse-regression based 
 methods and the CNN based models for the task.

\section{Discussion}
\label{sec:conclusion}
We have developed a non-negative least squares
based model reduction technique for discarding the
spurious association between the CNN
and n-gram representations to achieve an explainable model
for information extraction.
We have experimentally demonstrated that the reduced model 
on the n-gram presence features despite using
only a fraction of total text segments ($<10\%$) produce the same accuracy
as the CNN model.
Often lasso regression~\citep{tibshirani1996regression} is used in the
literature to render the sparse model.
However, our experiments with lasso regression failed to achieve an accurate
sparse model. 
We hypothesize that the lasso regression renders a relatively complex linear map 
that overfits the CNN representation by allowing the numerical cancellations 
in the modeling.  
By using the non-negativity constraints on the model, we were able to render a 
simpler sparse interpretable model.

Moving forward, we would like to examine the effectiveness of the model reduction 
on larger datasets.
Our future work would address any scalability hurdles that we may counter in solving large scale NNLS problems.
In this work, we employed a finite-sample variant of the integral Gaussian Process (SIGP)
primarily for rendering the class-label prediction. 
However, SIGP can be also useful to estimate the full predictive posterior distribution, 
which  may  yield a better uncertainty estimate than the softmax scores of a traditional CNN. 
SIGP admits a strictly larger set of functions than the Gaussian Process,
which includes solutions to Tikhonov regularization problem and Bayesian kernel models,
and thus can prove potentially useful for uncertainty quantification~\citep{tan2018subspace}
, which we would examine in our future work. 
%


\newpage

\bibliography{main}
\bibliographystyle{acl_natbib}



\end{document}